\documentclass[12pt]{article}
\usepackage{geometry}
\usepackage[utf8]{inputenc}
\usepackage[square,numbers, sort&compress]{natbib}
\usepackage{authblk}
\usepackage{amsmath,amssymb}
\usepackage{graphics}
\usepackage{graphicx}
\usepackage{verbatim}
\usepackage{algorithm}
\usepackage{algorithmic}

\usepackage[usenames,dvipsnames]{xcolor}

\usepackage{hyperref} 
\hypersetup{
colorlinks=true,       
linkcolor=OliveGreen,          
citecolor=Sepia,        
filecolor=magenta,      
urlcolor=NavyBlue,          
}

\usepackage[most]{tcolorbox}

\graphicspath{{figures/}}

\usepackage[normalem]{ulem}  
\newcommand{\jpi}[1]{{\color{Magenta} JPi: #1}}

\title{Chemoreception and chemotaxis of a three-sphere swimmer}

\author[1]{Stevens Paz}
\author[2]{Roberto F. Ausas}
\author[3]{Juan P. Carbajal}
\author[2]{Gustavo C. Buscaglia}
\affil[1]{Departamento de Matemáticas, Universidad del Valle, Calle 13 No. 100-00, Cali, Colombia.}
\affil[2]{Instituto de Ci\^encias Matem\'aticas e de Computa\c c\~ao, Universidade de S\~ao Paulo, S\~ao Carlos, Brazil.}
\affil[3]{Ostschweizer Fachhochschule, Institut für Energietechnik IET, Rapperswil, Switzerland. }

\begin{document}
\maketitle

\begin{abstract}
The coupled problem of hydrodynamics and solute transport for the Najafi-Golestanian three-sphere swimmer is studied, with the Reynolds number set to zero and P\'eclet numbers (Pe) ranging from 0.06 to 60. The adopted method is the numerical simulation of the problem with a finite element code based upon the FEniCS library.

For the swimmer executing the optimal locomotion gait, we report the Sherwood number as a function of Pe in homogeneous fluids and confirm that little gain in solute flux is achieved by swimming unless Pe is significantly larger than 10. 

We also consider the swimmer as an learning agent moving inside a fluid that has a concentration gradient. The outcomes of Q-learning processes show that learning locomotion (with the displacement as reward) is significantly easier than learning chemotaxis (with the increase of solute flux as reward). The chemotaxis problem, even at low Pe, has a varying environment that renders learning more difficult. Further, the learning difficulty increases severely with the Péclet number. The results demonstrate the challenges that natural and artificial swimmers need to overcome to migrate efficiently when exposed to chemical inhomogeneities. 

\end{abstract}

\section{Introduction}

Locomotion at microscopic scales is a fundamental process in several areas of biology, such as the study of the microbial life cycle, planktonic ecology, algal blooms, cell migration in tissue, fertilization and embryo development, and immune response. Liquid environments are the norm for microscopic creatures, which swim and crawl in many different and ingenious ways. Interest in the Fluid Mechanics of cell and micro-organism motion motivated early work by \citet{taylor1951analysis}, 
\citet{lighthill1976flagellar} 
and \citet{purcell1977life}, among others. Over the years, significant advances have been made that can be found in recent comprehensive reviews, such as those by \citet{guasto2012fluid}, \citet{elgeti2015physics}, 
\citet{lauga2016bacterial} and
\citet{goldstein2015green,goldstein2016batchelor}. 

In the last decade, interest in microscopic locomotion has expanded beyond natural processes toward artificial ones. In fact, the technology to deploy self-propelled micro-robots is becoming increasingly available or, at least, achievable 
\cite{becker2003self, peyer2012bacteria, gao2012cargo, pak2015theoretical, liao2019magnetically, tsang2020self}. 

The swimming strategies employed by natural and synthetic micro-organisms are highly diverse. To gain understanding of the basic mechanisms, theoretical studies have frequently relied on simplified models. Perhaps the simplest model is the \textit{squirmer}, which consists of a spheroidal particle with a given slip velocity over its surface and was introduced by \citet{lighthill1952squirming} and \citet{blake1971spherical} (see also \citet{stone1996propulsion}). Another simple early model is the Taylor sheet \cite{taylor1951analysis} (see also \citet{childress1981mechanics}), a flexible sheet that swims by deforming transversely according to a moving wave. 

In this study we address the three-sphere swimmer model, introduced by \citet{najafi2004simple, najafi2005propulsion} inspired by an earlier three-link mechanism proposed by \citet{purcell1977life}. This swimmer consists of three aligned spheres, with two contractible arms (of zero thickness) joining them (see Fig.~\ref{fig:ngsw}). It exhibits quite large surface motions because of the variable arm length with the simplicity of one-dimensional kinematics and spherical constituents. Because of its mathematical simplicity, the three-sphere swimmer has become a popular model \cite{golestanian2008three, golestanian2008analytic, alouges2009optimal, pickl2016fully, nasouri2019efficiency, berti2021numerical} 
with the additional advantage of having been realized experimentally \cite{grosjean2016realization}. 

Our focus is on the diffusive transport of dissolved species to (or from) the surface of the swimmer. This process is known as chemoreception. 
When the dissolved species is a nutrient, chemoreception governs the nutrient uptake of the organism and is central to its biological viability. The first studies \cite{berg1977physics, wiegel1983diffusion} 
considered a sphere in steady motion. Later, a spherical squirmer (self-propelled) was analyzed by \citet{magar2003nutrient} and \citet{magar2005average} 
and a suspension of spherical squirmers was simulated by \citet{ishikawa2016nutrient}. 

The chemoreception of kinematically more complex swimmers, such as the three-sphere swimmer model, has not been reported in the literature. With the increased interest in three-sphere swimmers and artificial swimmers in general, the first goal of this contribution is to make available a first study of the chemoreception of these swimmers for several swimming gaits and P\'eclet numbers. We adopt a numerical approach based on the finite element method which is an evolution of published work \cite{paz2020simulating, ausas2022finite}. 
Previous approaches have mainly relied on Oseen's approximation of the velocity field, which requires the three spheres to be very far apart. One of the few published similar works considers a biflagellated sphere \cite{tam2011optimal}, 
with the flagella moving according to the swimming gates of the alga Chlamydomonas. 

Our second goal concerns chemotaxis, that is, the ability of the swimmer to move towards regions of higher solute (nutrient) concentration based on chemoreception data. Recent attempts have shown that artificial swimmers can learn locomotion by reinforcement learning algorithms \cite{tsang2020self, misra2020kinematic, berti2021numerical, alageshan2020machine, qiu2022navigation, muinos2021reinforcement}. 
Learning chemotaxis was considered by \citet{hartl2021microswimmers}, 
who applied a genetic learning algorithm to the three-sphere swimmer and showed that it can successfully navigate static and time-dependent chemical environments. However, they considered a simplified fluid dynamical model that did not account for the swimmer's impact on the concentration field. In this contribution we report on some learning experiments with realistic interaction of the swimmer with the environment. A basic Q-learning algorithm \cite{watkins1992q} 
is used to process data from random sequences of actions. Then, from looking at the learning outcomes, the increased difficulty of learning chemotaxis over learning locomotion becomes apparent. The difficulty grows severely for Péclet numbers above 1.

\section{Governing equations}

\begin{figure}[t]
\begin{center}
\includegraphics[width=.7\textwidth]{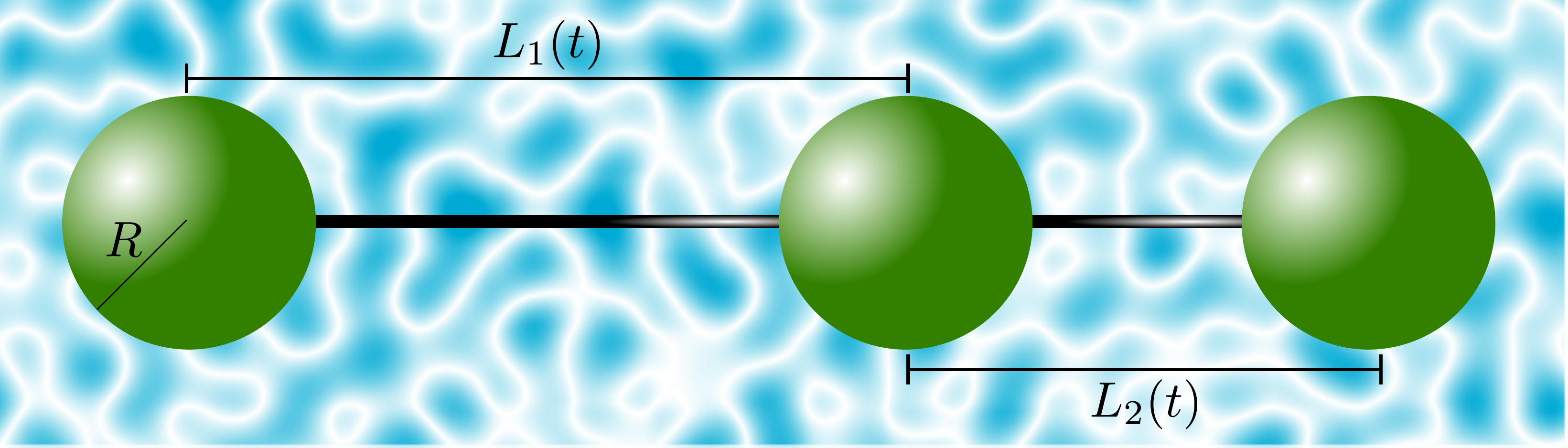}
\end{center}
\caption{Three-spheres swimmer considered by \citet{najafi2004simple}.}\label{fig:ngsw}
\end{figure}

We consider the one-dimensional Najafi-Golestanian swimmer \cite{najafi2004simple}, which consists of three spheres of radius $R$ joined by two extensible virtual (zero thickness) arms (see Fig.~\ref{fig:ngsw}). The variable lengths $L_1(t)$ and $L_2(t)$ of the arms, which are given functions of time, produce net displacements (in the laboratory frame) $X_1(t)$, $X_2(t)$ and $X_3(t)$ of the centers of the spheres and thus of the center of mass of the swimmer, $X(t)$. Notice that, since $X=(X_1+X_2+X_3)/3$,
$$
X_1=X-\frac{2L_1}{3}-\frac{L_2}{3},\quad
X_2=X+\frac{L_1}{3}-\frac{L_2}{3},\quad
X_3=X+\frac{L_1}{3}+\frac{2L_2}{3}~.
$$
The centers $\mathbf{C}_i$ ($i=1,2,3$) of the spheres are assumed lying along the $x_1$ axis, i.e., $\mathbf{C}_i = (X_i,0,0)^T$. The domain occupied by the solid spheres is thus 
$$
\mathcal{B}(X,L_1,L_2) =  \mathcal{B}_1 \cup \mathcal{B}_2 \cup \mathcal{B}_3~,\qquad \text{with } \quad \mathcal{B}_i =
\left \{ \mathbf{x}\in\mathbb{R}^3~|~ \|\mathbf{x}-\mathbf{C}_i\| \leq R \right \}
$$
The swimmer evolves inside a large domain $\Omega$ with boundary conditions mimicking an infinite medium. The fluid domain is $\Omega_{f}\left(X,L_1,L_2\right) = \Omega\setminus\mathcal{B}\left(X,L_1,L_2\right)$, filled with an inertialess incompressible Newtonian fluid of viscosity $\mu$. The Cauchy stress of the fluid is thus given by
\begin{equation}
    \boldsymbol{\sigma} = -p\,\mathbf{I} + \mu\,\left (\nabla \mathbf{u} + \nabla \mathbf{u}^T \right )~,
\end{equation}
where $p$ and $\mathbf{u}$ are the fluid's pressure and velocity fields, respectively, and $\mathbf{I}$ is the identity matrix. 

Assuming no external forces acting on the swimmer, the {\bf hydrodynamic problem} reads:

\bigskip

{\em Given $X(0)$, $L_1(t)$ and $L_2(t)$, for $t>0$ find $X(t)$, $\mathbf{u}(\mathbf{x},t)$ and $p(\mathbf{x},t)$ such that
\begin{eqnarray}
\nabla \cdot \boldsymbol{\sigma} & = & 0~,\qquad \text{in } \Omega_f(t) \label{eq:dif1}\\
\nabla \cdot \mathbf{u} & = & 0~,\qquad \text{in } \Omega_f(t) \label{eq:dif2}\\
\int_{\partial \mathcal{B}(t)} \boldsymbol{\sigma}\cdot \mathbf{\check{n}} ~dS & = & 0~, \label{eq:dif3}\\
\mathbf{u} & = & \left(\frac{dX}{dt} -\frac23 \frac{dL_1}{dt} -\frac13 \frac{dL_2}{dt}\right)\hat{\mathbf{e}}~,\qquad \text{on } \partial \mathcal{B}_1  \label{eq:dif4}\\
\mathbf{u} & = & \left(\frac{dX}{dt} +\frac13 \frac{dL_1}{dt} -\frac13 \frac{dL_2}{dt}\right)\hat{\mathbf{e}}~,\qquad \text{on } \partial \mathcal{B}_2  \label{eq:dif5}\\
\mathbf{u} & = & \left(\frac{dX}{dt} +\frac13 \frac{dL_1}{dt} +\frac23 \frac{dL_2}{dt}\right)\hat{\mathbf{e}}~,\qquad \text{on } \partial \mathcal{B}_3  \label{eq:dif6}
\end{eqnarray}
}

Equations (\ref{eq:dif1})-(\ref{eq:dif2}) are the well known Stokes equations, (\ref{eq:dif3}) is the force-free condition where $\mathbf{\check{n}}$ is the unit normal outwards of $\Omega_f$, and (\ref{eq:dif4})-(\ref{eq:dif6}) impose the no-slip condition at the surface of each sphere, being $\hat{\mathbf{e}}$ the unit positive direction of motion.

Coupled with the hydrodynamic problem is the {\bf solute transport problem}, in which the unknown is the concentration field $C(\mathbf{x},t)$. It reads:

\bigskip

{\em Given the initial concentration field $C(\mathbf{x},t=0)$, for $t>0$ find $C(\mathbf{x},t)$ satisfying
\begin{eqnarray}
\frac{\partial C}{\partial t} + \mathbf{u}\cdot \nabla C - D\,\nabla^2 C & = & 0~,\qquad \qquad \text{in } \Omega_f~, \label{eq:trans1}\\
D \nabla C \cdot \mathbf{\check{n}} &=& -\phi_i\, C~,\qquad \text{on } \partial \mathcal{B}_i~. \label{eq:trans2}
\end{eqnarray}
with boundary conditions $C = C_\infty$ far away from the swimmer.
}

In (\ref{eq:trans1}), $D$ is the diffusivity of the solute.
The so-called Robin boundary condition (\ref{eq:trans2}) contains the parameters 
which govern the permeability of the spheres' surfaces to the solute. As $\phi_i \to \infty$, the Robin condition tends to the Dirichlet condition $C=0$. Of crucial importance in our analysis is the total solute flux $J$ to the swimmer, defined as
\begin{equation}
    J(t) = \int_{\partial \mathcal{B}(t)} D \nabla C \cdot \mathbf{\check{n}}~dS~.
\end{equation}

\section{Numerical method: Description and validation}

The numerical method is an adaptation of the Arbitrary Lagrangian-Eulerian (ALE) finite element methods discussed in previous work by the authors in the context of squirmers \cite{paz2020simulating} 
and of flexible swimmers \cite{ausas2022finite}. 
The current implementation relies heavily on the FEniCS package \cite{AlnaesBlechta2015a}. 

The spatial interpolation of velocity and pressure adopts $P_1$ conforming finite elements, stabilized by the well-known ASGS (or GLS) technique \cite{hughes1986new, hughes1987new, codina2002stabilized}. 
The interpolation of the concentration field is carried out with $P_2$ finite elements. Stabilization of advection is also performed by the ASGS method. Large deformations of the mesh are mitigated by elastic mesh updates plus frequent remeshing based on a mesh quality criterion.

The time stepping strategy follows a second-order predictor-corrector method for the hydrodynamic problem and a second-order trapezoidal scheme for the solute transport. More details can be found in Appendix A.

\begin{figure}[t]
\begin{center}
\includegraphics[width=.7\textwidth]{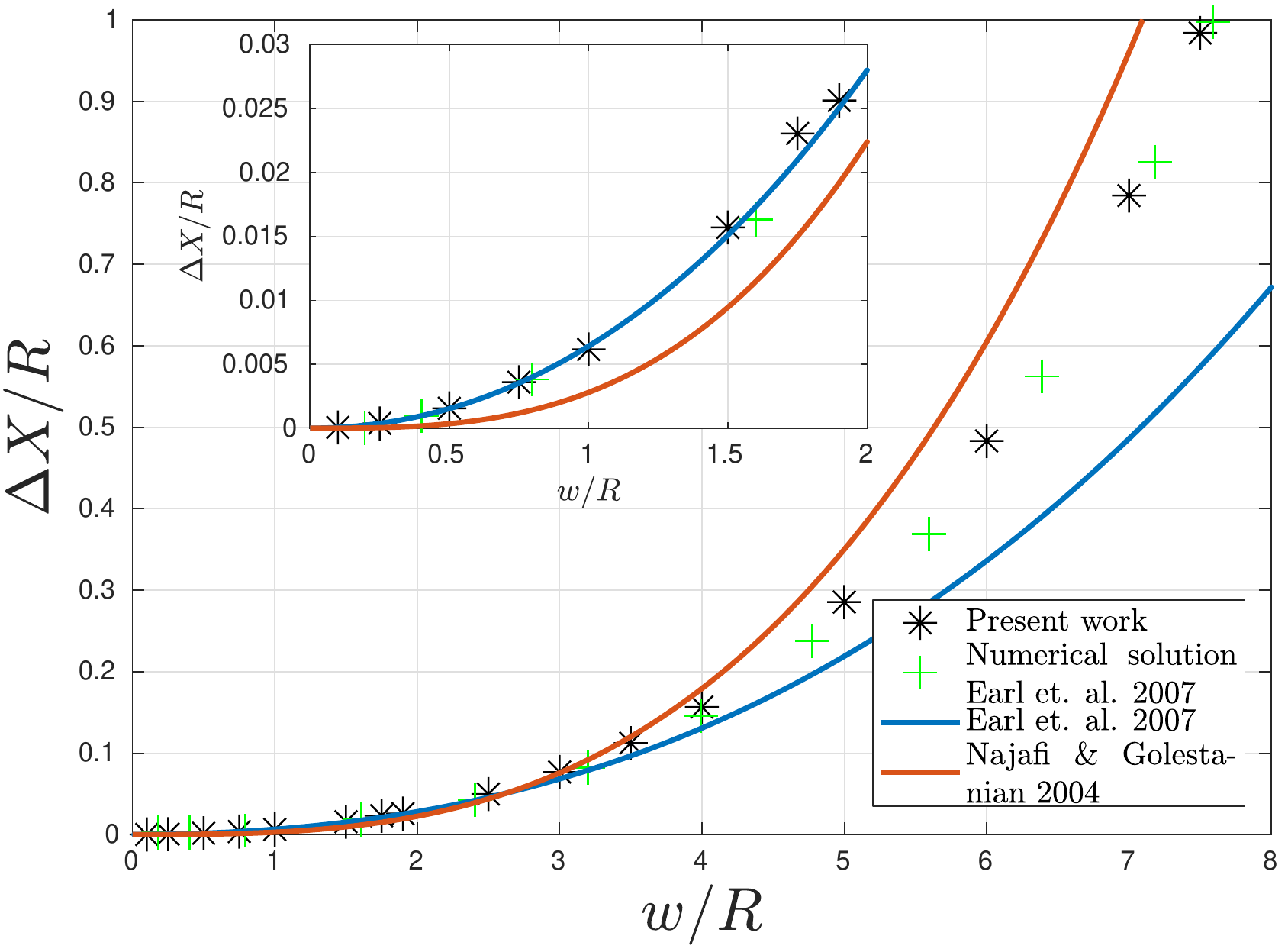}
\end{center}
\caption{Asymptotic behaviour of normalized displacement $\Delta X/R$ when $w/R\to 0$ for $3$-linked spheres swimmer. The $P1/P1$-stabilized solution (asterisk marker) is compared against the formulae from \citet{earl2007modeling} and \citet{najafi2004simple}. 
}\label{fig:cmpdisp}
\end{figure}

Several 
tests were performed to validate the numerical code:

\begin{description}
\item[a)] Displacement achieved by the swimming gait of the Najafi-Golestanian swimmer. By modifying the length of one arm at a time, between the maximum length $W$ and the minimum one $W-w$, the swimmer advances a distance $\Delta X$ per gait. 
Previous asymptotic and numerical estimates \cite{najafi2004simple, earl2007modeling} 
are compared in Fig.~\ref{fig:cmpdisp} to our numerically obtained values, showing excellent agreement. 
\item[b)] Chemoreception of a sphere towed at constant velocity in a fluid. It is known that for a towed sphere the solute flux $J$, for Dirichlet conditions ($\phi \to +\infty$), is well approximated by the expression proposed by \citet{clift1978bubbles} and \citet{karp1996nutrient}
\begin{equation}
    \text{Sh} = \frac{J}{J_0} = \frac12 \left ( 1 + \left ( 1 + 2 \text{Pe} \right )^{\frac13} \right )~, \label{eq:clift}
\end{equation}
where $J_0$ is the solute flux when towing velocity $V$ is zero (purely diffusive transport) and Pe $=VR/D$ is the P\'eclet number. The ratio $J/J_0$ is known as {\em Sherwood number} Sh. Our numerical results were verified to agree within 3\% with Clift's formula in the whole range $10^{-2}\leq \text{Pe} \leq 10^2$.
\end{description}

\section{Solute flux toward a three-sphere swimmer}
 
\begin{figure}[t]
\begin{center}
\includegraphics[width=.7\textwidth]{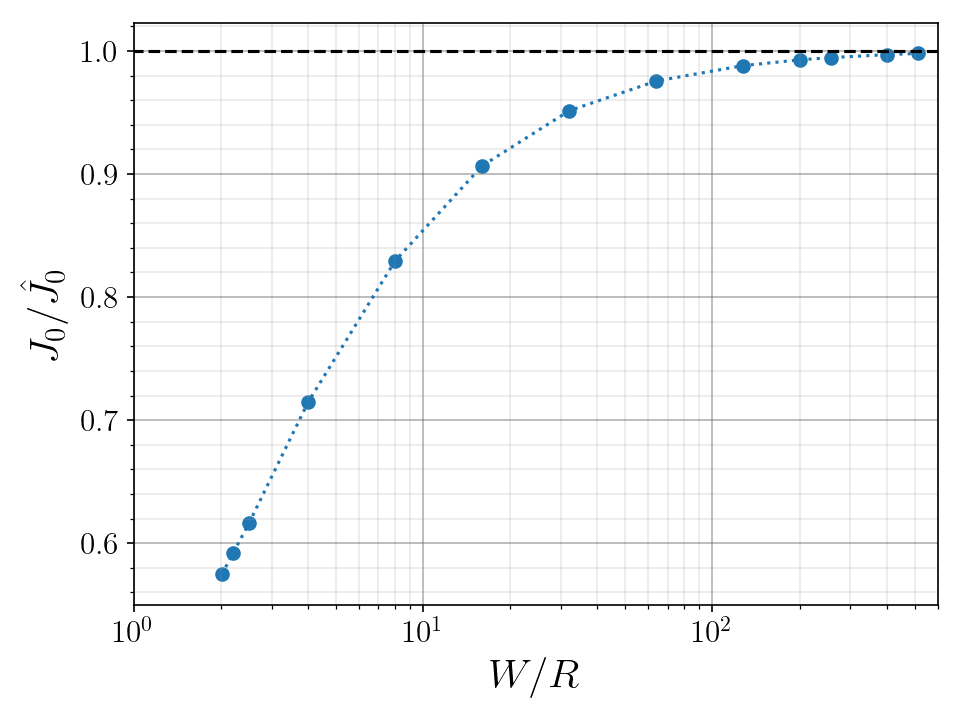}
\end{center}
\caption{$J_{0}/\hat{J}_{0}$ as a function of $W/R$.}\label{fig:WRvsJ0sp}
\end{figure}

The baseline solute flux $J_0$ of a three-sphere swimmer is defined here as the steady diffusive flux at its surface when the arms of the swimmer are totally extended. In what follows we concentrate on the Dirichlet condition case, i.e., all $\phi_i \to +\infty$. If $W\gg R$ the analytic solution $C(r)=C_\infty(1-\frac{R}{r})$ for a sphere yields the baseline flux $\hat{J}_0 = 12 \pi R D C_\infty$. For smaller values of $W/R$ one gets smaller values of $J_0$, as shown in Fig.~\ref{fig:WRvsJ0sp}. 

\begin{figure}[t]
\begin{center}
\includegraphics[width=.5\textwidth]{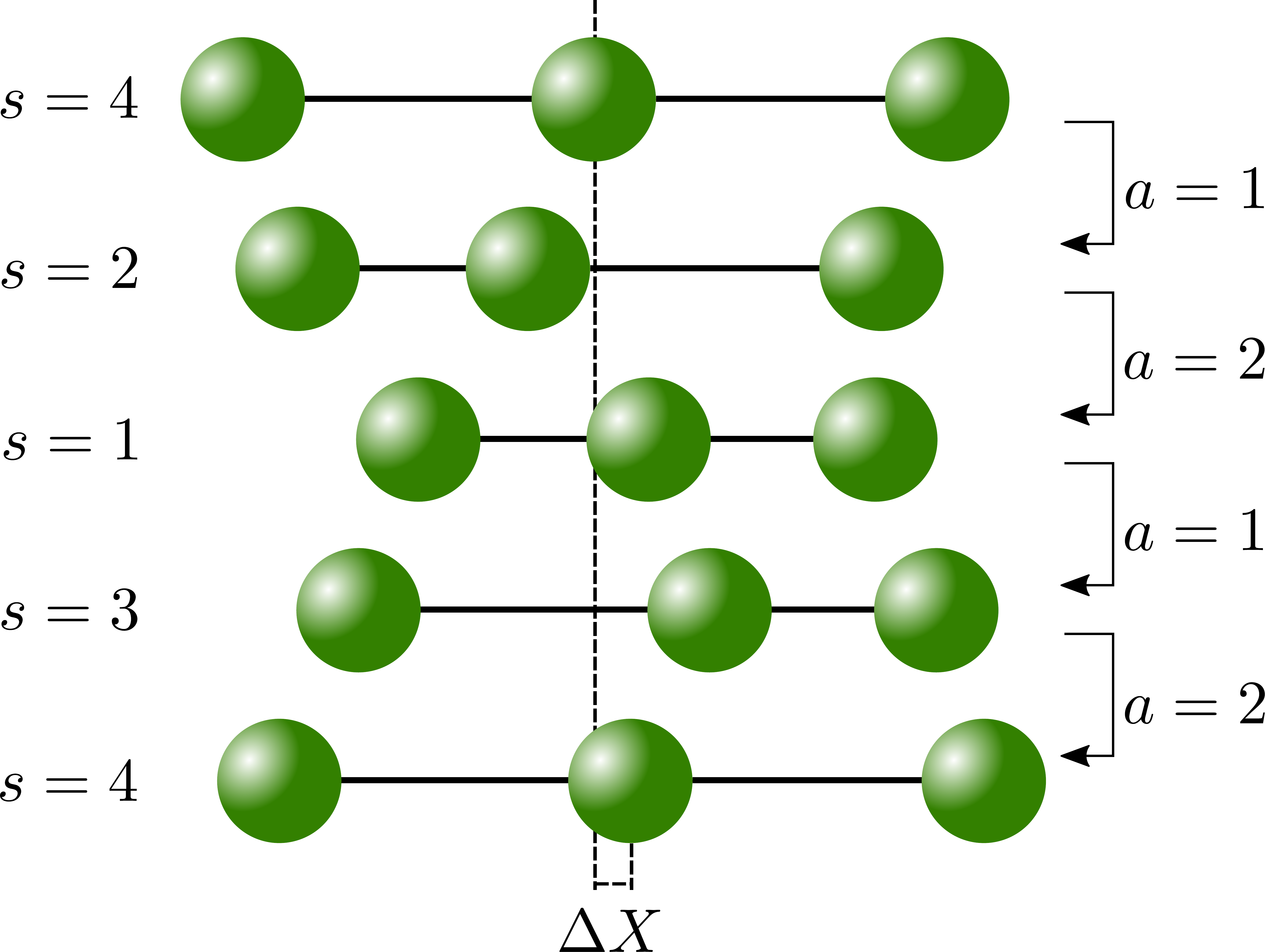}
\end{center}
\caption{Swimming gait for the three-spheres swimmer considered by \citet{najafi2004simple}. The sequence is from top to bottom and follows the numeration introduced in Sec. \ref{ssec:RLF} for states ($s$) and actions ($a$).}\label{fig:swgait}
\end{figure}

\begin{figure}[t]
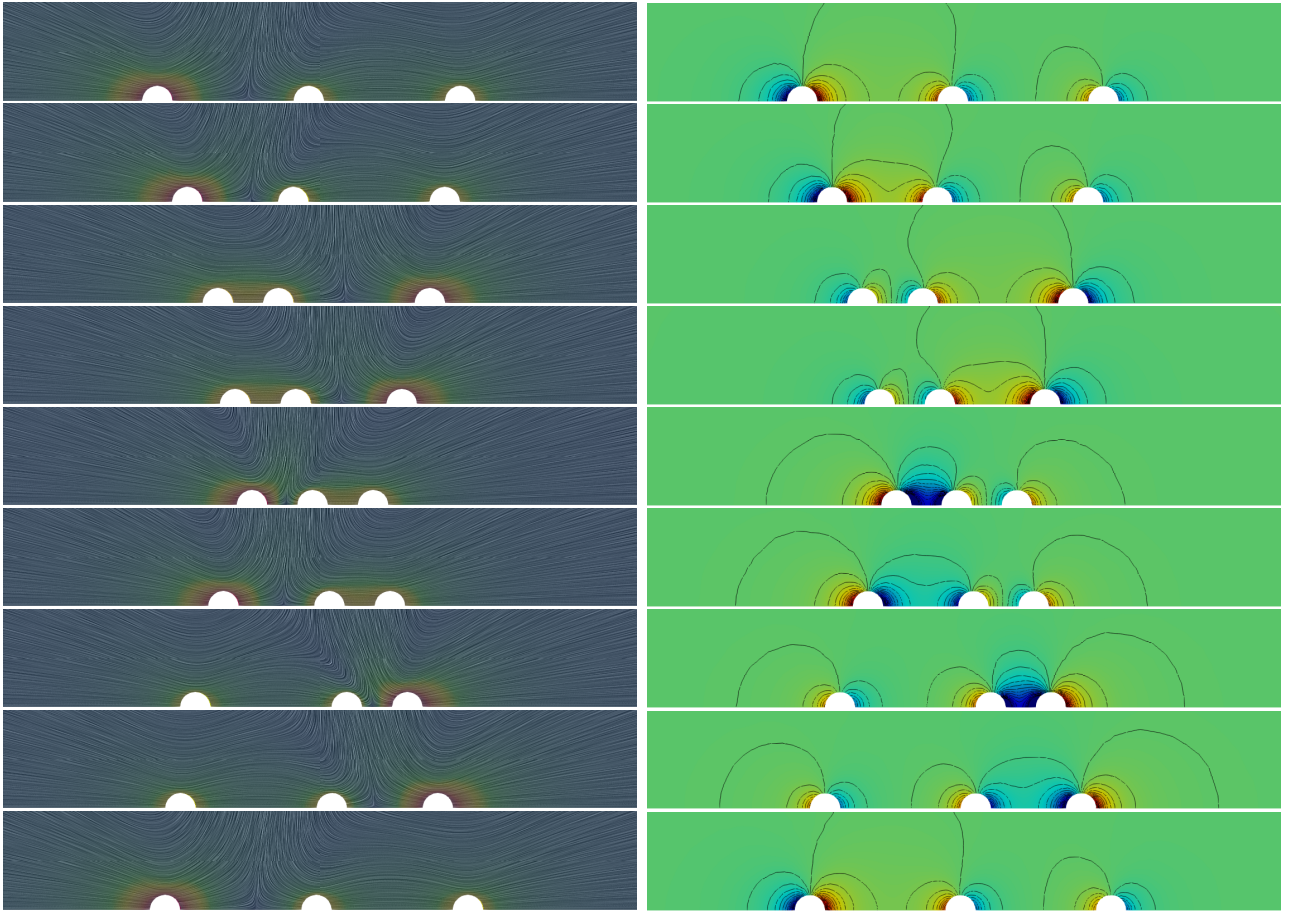

\begin{center}
\foreach \x in {0000,0005,0010,0015,0020,0025,0030,0035,0040}
{
\includegraphics[width=.49\textwidth]{vD1.\x.png}
\includegraphics[width=.49\textwidth]{PD1.\x.png}\par
}
\end{center}
\caption{Swimming gait for the three-spheres swimmer considered by \citet{najafi2004simple}. Left: velocity magnitude and streamlines. Right: pressure field and contour lines. In both cases, colors going from red to blue indicating maximum to minimum quantity value.}\label{fig:vpevol}
\end{figure}

In this study we consider the basic non-reciprocal {\em swimming gait} discussed by \citet{najafi2004simple}, by which the arms move one at a time, taking turns (i.e., starting from the extended position, contract arm 1, contract arm 2, extend arm 1, extend arm 2, repeat) as shown in Fig.~\ref{fig:swgait} and Fig.~\ref{fig:vpevol}. An equivalent gait is obtained by reversing the sequence in time, the only difference being that the direction of the movement is to the left.

When it's time for one specific arm to move, it goes to complete extension ($L_i=W$) to complete contraction ($L_i=W-w$), if it was extended, or viceversa, if it was contracted. Though the net displacement obtained by executing the gait does not depend on the speed at which the sequence of movements is made (because fluid inertia is negligible), the solute transport is affected by the speed. The simplest case was adopted, in which all arm extensions and contractions are performed at uniform, constant speed $S$. 
The duration of the gait is thus $T=4w/ S$, after which the swimmer recovers its initial posture, but its position is affected by a translation $\Delta X$. The computation of $\Delta X$ requires solving the {\em hydrodynamic problem} (\ref{eq:dif1})-(\ref{eq:dif6}).
The values of $\Delta X$ obtained with our numerical code are compared to reference ones published by Earl et al (2007) and to asymptotic approximations in Fig.~\ref{fig:cmpdisp}.
The good agreement is an additional validation of the code used in this work. 
The average velocity of the center of mass of the swimmer is $V=S \Delta X / (4w)$. Let us denote $\Delta \mathbf{X}=(\Delta X,0,0)^T$. The solution of the hydrodynamic problem is periodic in the swimmer's frame and thus only needs to be solved in time along one gait duration (say, between $t=0$ and $t=T$), since
\begin{eqnarray}
X(t+T) & = & X(t) + \Delta X~, \\
\mathbf{u}(\mathbf{x}+\Delta \mathbf{X},t+T) & = & \mathbf{u}(\mathbf{x},t)~, \\
p(\mathbf{x}+\Delta \mathbf{X},t+T) & = & p(\mathbf{x},t)~.
\end{eqnarray}
The {\em transport problem (\ref{eq:trans1})-(\ref{eq:trans2})}, on the other hand, depends upon the initial condition and evolves in time as the immediate surroundings of the swimmer are gradually depleted and the boundary layers produced by the swimmer's movement develop.
Eventually, after some development time $\mathcal{T}$, the concentration field also reaches a periodic evolution in the swimmer's frame, namely, for $t > \mathcal{T}$, 
\begin{equation}
    C(\mathbf{x}+\Delta \mathbf{X},t+T)  =  C(\mathbf{x},t)~.
\end{equation}
This implies that $J(t)$ is also periodic, $J(t+T)=J(t)$. Our aim is to provide data about the {\em average solute flux}
\begin{equation}
    \overline{J}=\frac{1}{T} \int_{t_0}^{t_0+T} J(t)~dt~
\end{equation}
once periodic conditions are attained, i.e., $t_0 > \mathcal{T}$. The average solute flux {\em when the swimmer is executing the swimming gait} is denoted by $J_s(\text{Pe})$.

\noindent{\bf Remark:} The system necessarily arrives at a periodic state because the hydrodynamic problem is totally decoupled from the transport problem, the latter being linear and parabolic (albeit in a periodically-varying domain). Nonlinearities by which the concentration field affects the hydrodynamics may lead to more complex behaviors.

The parameters of the problem are $R$, $W$, $w$, $D$ and $S$. The far field concentration $C_\infty$ is irrelevant, taken as unity. The relevant non-dimensional variables are $W/R$, $w/R$ and the P\'eclet number
\begin{equation}
    \text{Pe} = \frac{SR}{D} ~.
\end{equation}
We have considered $W/R=10$ combined with different values of $w/R = 2, 4$ and $6$. Notice that in the last case the spheres are just one diameter apart when the arm is in the contracted position.

\begin{figure}[t]
\begin{center}
\includegraphics[width=.7\textwidth]{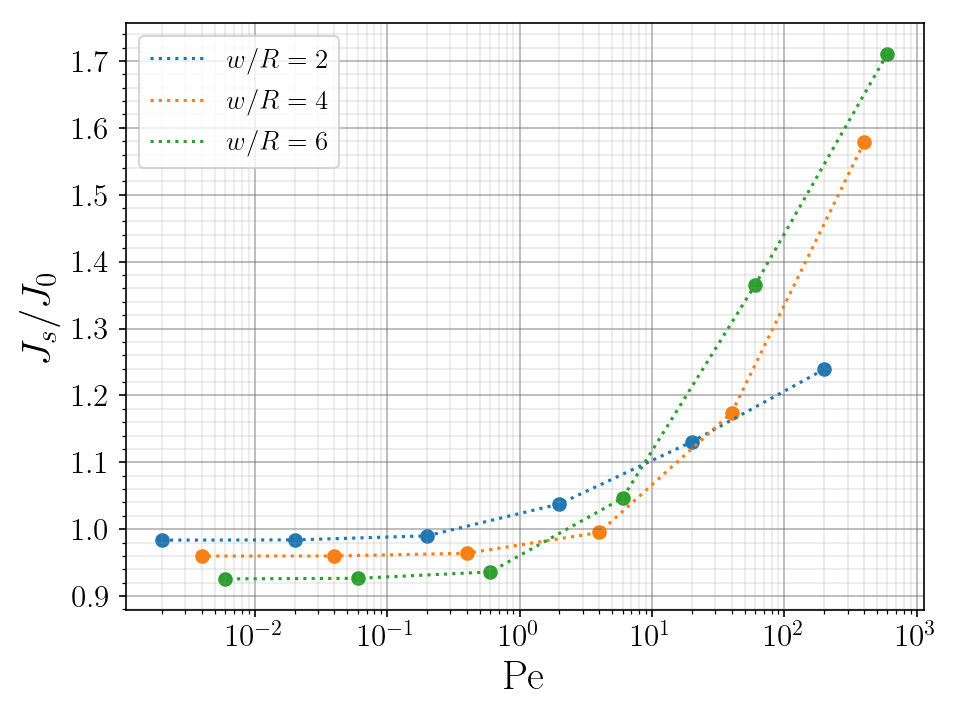}
\end{center}
\caption{Sherwood number as a function of Pe for $w/R = 2, 4$ and $6$.}\label{fig:peshcmpwR}
\end{figure}

The solution of the full problem (hydrodynamics+transport) yields the average Sherwood number $J_s/J_0$ as a function of Pe for each case, as shown in Fig.~\ref{fig:peshcmpwR}. It is worth pointing out that for very low Pe the solute flux is maximal when the swimmer is immobile with both arms extended. The swimming gait takes the arms out of the optimal position and thus results in $J_s<J_0$, as evinced by Fig. \ref{fig:peshcmpwR} for Pe $< 5$.

\begin{figure}[t]
\begin{center}
\includegraphics[width=.48\textwidth]{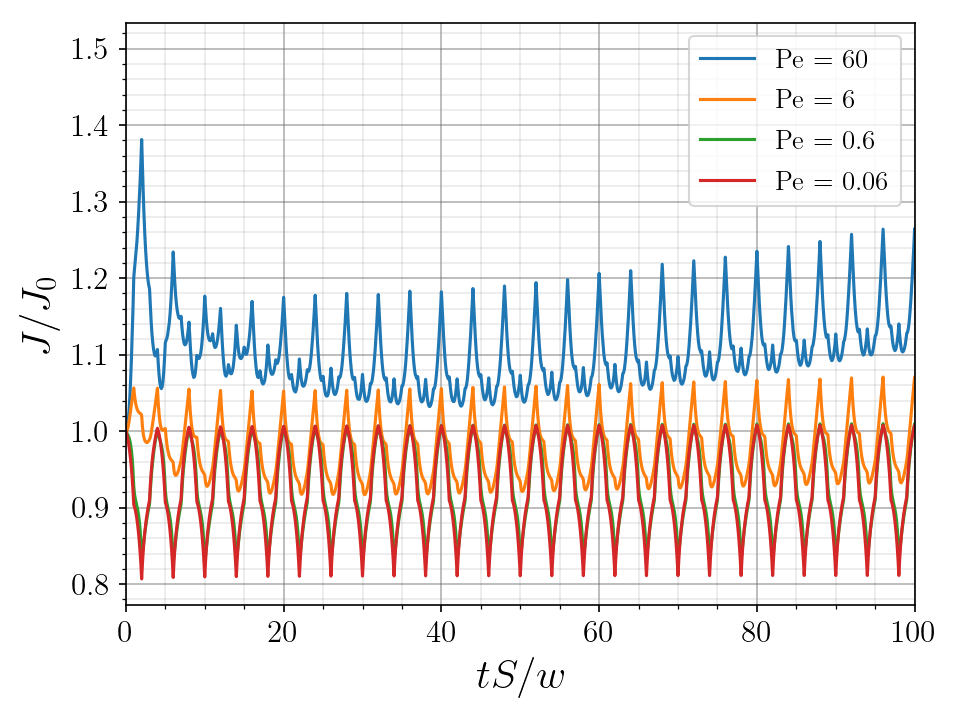}
\includegraphics[width=.48\textwidth]{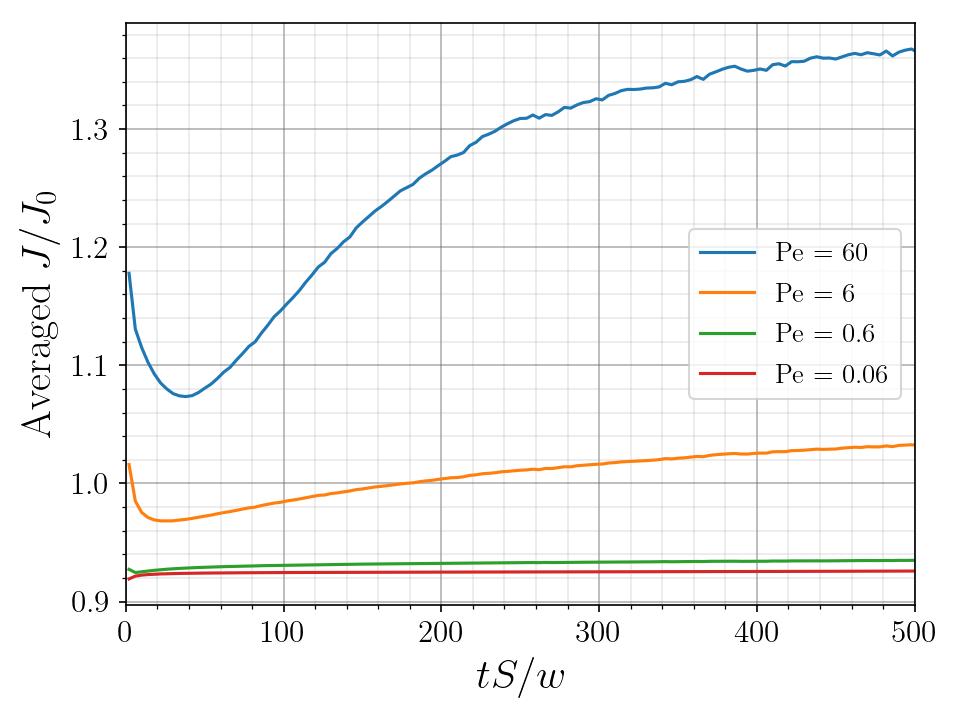}
\end{center}
\caption{Left: Time histories of $J(t)$ up to non-dimensional time $t=100$ after starting the swimming gait at different Péclet numbers. Right: Long term behavior of the flux averaged over the swimming gait (four actions) up to time $t=500$.}\label{fig:petrans}
\end{figure}

\begin{figure}[t]
\begin{center}
$tS/w = 0$\par
\includegraphics[width=.79\textwidth]{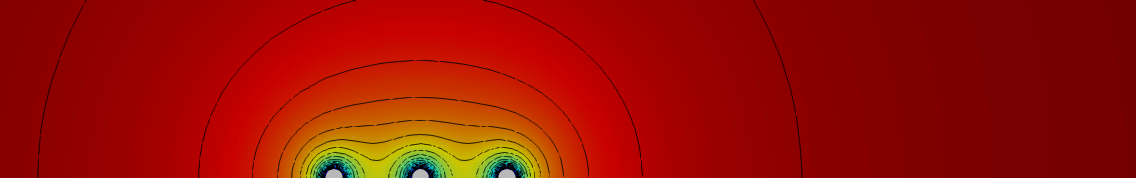}\par
$tS/w = 240$\par
\includegraphics[width=.79\textwidth]{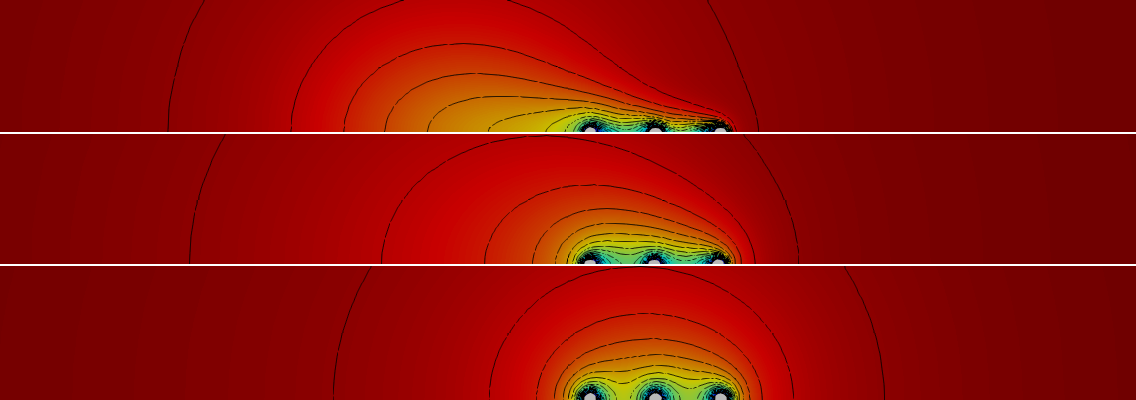}\par 
$tS/w = 500$\par
\includegraphics[width=.79\textwidth]{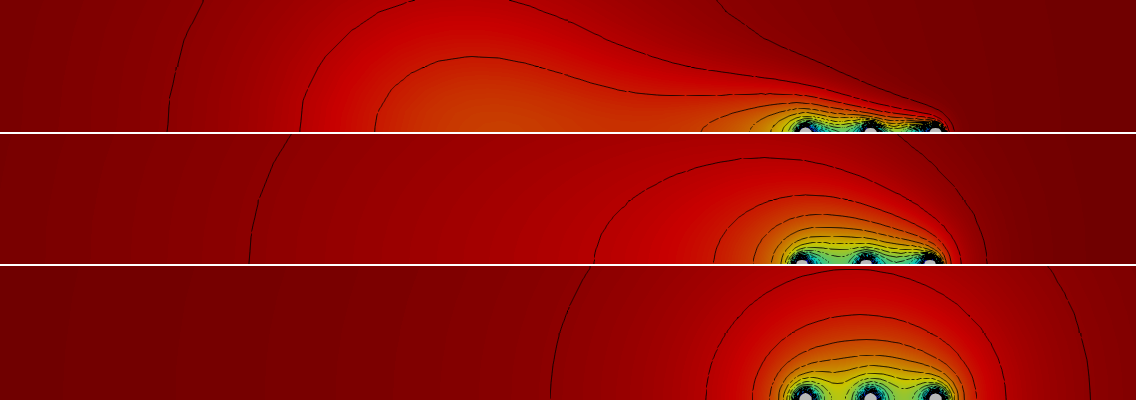}
\end{center}
\caption{Snapshots of the concentration wake for different times and P\'eclet numbers (contour lines in black color and colors going from red to blue indicating maximum to minimum concentration). Top: initial condition for all cases. Center: time $tS/w = 240$ and Pe $ = 60, 6, 0.6$ (top to bottom, respectively). Bottom: time $tS/w =500$ and Pe $ = 60, 6, 0.6$ (top to bottom, respectively)}\label{fig:cfevol}
\end{figure}

In Fig.~\ref{fig:petrans} we present the transient fluxes $J(t)/J0$ starting from an initial condition that is the steady-state solution corresponding to the initial posture of the swimmer. We include, in Fig.~\ref{fig:cfevol}, some sample snapshots of the concentration wake of the swimmer at different times.

\section{Q-Learning chemotaxis}

\subsection{The reinforcement learning framework}\label{ssec:RLF}

One can view the swimmer as an {\em agent} in discrete time performing a {\em decision process} \cite{fan2020theoretical}. The time unit is $w/S$, the time it takes to contract an extended arm or vice versa. In its simplest formulation, the agent has four possible {\em states}: both arms contracted (state 1), arm 1 contracted and arm 2 extended (state 2), arm 1 extended and arm 2 contracted (state 3) and both arms extended (state 4). The sequence of states that corresponds to the swimming gait (toward the right, i.e. $x_1 > 0$) 
is 1-3-4-2-1-3-$\ldots$. The immobile extended mode is a sequence 4-4-$\ldots$. 

We define that the agent {\em must move exactly one arm in each time interval}, from the fully extended position to the fully contracted position or vice versa. The only decision to be taken at each discrete instant $t$ is whether to move arm 1 (action 1) or arm 2 (action 2). At each of the four possible states there are just two possible actions. This kinematic restriction on the agent rules out the possibility of the swimmer remaining still, in any position. The swimming gait to the right, starting at state 4, corresponds to the sequence of actions 1-2-1-2-$\ldots$ .  

The agent defined above can be made to follow a {\em policy}, i.e., a rule that to each state $s_t\in \{1,2,3,4\}$ associates a probability $\pi_t(s_t,a=1)$ of taking action $a=1$ at time $t$ (and thus $\pi_t(s_t,a=2)=1-\pi_t(s_t,a=1)$ of taking action $a=2$). We only consider here time-independent policies, so that we drop the subscript $t$ from $\pi_t$ hereafter. A totally random policy selects, at each decision time, action 1 or 2 randomly with 50\% probability each (i.e., $\pi(s_t,1)=\pi(s_t,2)=0.5$). The swim-to-the-right gait is a totally deterministic policy that takes action 1 if $s_t=1$ or $s_t=4$ and takes action 2 if $s_t=2$ or $s_t=3$. This corresponds to $\pi(1,1)=\pi(4,1)=1$ and $\pi(2,1)=\pi(3,1)=0$. This particular policy is denoted by $\pi^*$. Other policies are of course possible, and the identification of a convenient policy is referred to as {\em learning}.

Whichever the policy, each action causes the agent to interact with the fluid. Assume action $a_t$ is taken at time $t$, when the agent is at state $s_t$ in an environment described by the variable $\theta_t$. After the action, the state of the agent is altered to another state $s_{t+1}$ and the environment changes to $\theta_{t+1}$. In our setting, there is only one $s_{t+1}$ possible as a consequence of each action, and it is independent of the environment. 

The environment variable $\theta_t$ represents all the relevant information about the environment at time $t$. To define it more precisely, we consider our agents moving inside a domain $\Omega$ that is fixed in the laboratory frame. The geometry of the fluid environment (the domain $\Omega_f$ discussed in Section 2) is fully determined, at time $t$, by $X_t=X(t)$, $L_1(t)$ and $L_2(t)$. The environment variable here is $X_t$, since $L_1(t)$ and $L_2(t)$ are state variables. Looking now at the solute transport problem, the only additional information that affects the evolution of the system between $t$ and $t+1$ is the initial concentration field $C_t=C(\mathbf{x},t)$ (with $\mathbf{x}$ arbitrary in $\Omega_f(t)$). The environment variable $\theta_t$ is thus defined as $\theta_t = (X_t,C_t)$. By solving the hydrodynamic+transport problem from time $t$ to time $t+1$ one can update the environment variables. Formally, we write this as
$$
\theta_{t+1} = \Theta(s_t,a_t,\theta_t)~,
$$
where the update function $\Theta$ is not explicitly known. 

The learning problem we consider is that of finding a policy $\pi^*$ that maximizes some additive {\em total reward} 
\begin{equation}
    R(\pi)=\sum_{\tau > 0} \gamma^{\tau} r_{\tau}~, 
\end{equation}
where $\gamma\in (0,1)$ is the discount factor and $r_{\tau+1}$ is the {\em instantaneous reward} obtained by performing the action $a_{\tau}$ selected by the policy $\pi$. As $\gamma$ increases toward 1, the total reward gives more importance to the long term consequences of current actions.

\subsection{Rewards associated to chemotaxis}

Consider the 3-sphere agent immersed in a fluid where there exists a constant gradient of solute concentration, i.e., far away from the agent the concentration field is given by $C(x,y,z,t)= g\,x$, where $g>0$. Assume that the agent wants to maximize its access to the solute, then it is easily seen that its optimal policy is nothing but the swimming gait (to the right). In terms of an additive reward this can be set in several ways, of which we consider the following three:
\begin{description}
\item[R1)] Displacement reward: In this case the instantaneous reward is simply
\begin{equation}
    r_{t+1} = X_{t+1} - X_{t}~,
\end{equation}
so that $\sum_{t=1}^{N}  r_t = X(N)-X(0)$. Notice that this reward is purely hydrodynamic (it is independent of the concentration field). It corresponds to "learning to swim", or "learning locomotion", rather than "learning chemotaxis". \citet{tsang2020self} have shown that the Q-learning algorithm \cite{watkins1992q} successfully identifies optimal policies for R1. They used a simplified hydrodynamic model (Oseen approximation) that is only valid when $W-w \gg R$. Our hydrodynamic model (\ref{eq:dif1})-(\ref{eq:dif6}), being exact, is not subject to this condition. In fact, we show below that Q-learning is also very efficient at learning to swim when $W-w = 4R$. 

The rewards we consider next are true chemotactic rewards and, unlike R1, they are observable to the agent without knowing its relative position, w.r.t. the lab. frame. 
Our aim is to evaluate the difficulty of learning chemotaxis as compared to that of learning locomotion.

\item[R2)] Intake accumulation: This reward is the total intake of solute during the interval $[t,t+1]$, 
\begin{equation}
    r_{t+1}= \int_t^{t+1} J(\tau)~d\tau~,
\end{equation}
so that $\sum_{t=1}^N r_t = \int_0^N J(\tau)~d\tau$.
It expresses the drive to maximize the total intake along the evolution of the system.
\item[R3)] Intake increase: This reward is defined as
\begin{equation}
    r_{t+1}= J(t+1)-J(t)~,
\end{equation}
so that $\sum_{t=1}^{N}  r_t = J(N)-J(0)$. In drives the agent towards maximizing the solute intake rate at the end of the evolution.
\end{description}

The consideration of the solute transport variables in rewards R2 and R3 increases the complexity of the agent-environment problem, because $C(\cdot,t)$ is a {\em field} and thus high-dimensional. 

\subsection{The Q-learning algorithm}

Q-learning is based on the value-iteration algorithm, which provably (under suitable assumptions) converges to a solution of the Bellman optimality equation.

The algorithm successively updates an {\em action-value matrix} $Q(s,a)$ which represents the maximum total reward that can be expected if action $a$ is taken when the agent is at state $s$.

An {\em experience} is defined as a 4-tuple $e_{t+1}=(s_t,a_t,s_{t+1},r_{t+1})$ consisting of the agent's state prior to the action, the action taken, the agent's state after the action, and the instantaneous reward collected. Given a set of experiences $e_1, e_2, \ldots, e_N$, and initializing $Q_0$ with zeroes, $Q_{t}$ is updated as
\begin{equation}
     Q_{t+1}(s_t,a_t) = (1-\alpha) Q_t(s_t,a_t) + \alpha\,\left (
     r_{t+1} + \gamma\,\max_{a\in\{1,2\}} Q_t(s_{t+1},a) \right )~,
 \end{equation}
 where $\alpha$ is the so-called {\em learning rate}. After the $N$ updates, one obtains matrix $Q_N$. The {\em learned policy} $\pi_{\text{\tiny QL}}$ is totally greedy in $Q_N$, i.e.,
 $$
 \pi_{\text{\tiny QL}}(s,a)=0\quad \forall (s,a)~\mbox{s.t.}~
 Q_N(s,a) < \max_{a'} Q_N(s,a')~.
 $$
 For each set of experiences, the learned policy $\pi_{\text{\tiny QL}}$ depends on the two parameters $\gamma$ and $\alpha$. 
 
 Q-learning is an {\em off-policy} algorithm, in the sense that the set of experiences can be taken from any evolution of the system that repeatedly visits all possible states.
 We generate the experience sets with random policies, as will be described later on.
 Along the generation of each set we store rewards R1, R2 and R3 for each experience.
 This allows us to compare the learning outcomes of the Q-learning algorithm for all rewards and different environments (Pe number) over the same experience set.

\subsection{Numerical experiments}

We consider an agent with geometrical parameters $W/R=10$ and $w/R=6$ that is immobile at state $s_0$ and center-of-mass position $X_0$ for a long time in a homogeneous environment. The concentration field has reached, at time $t=0$, its steady state with value $C_\infty(x,y,z)=g\,x$ far from the agent and 0 at its surface. At $t=0$ we start the decision process following a totally random policy and solve the hydrodynamic+transport problem along $N$ actions until time $t=N$, storing the sequence of experiences corresponding to each reward. The solution is computed numerically with the FEniCS-based solver, using a time step $\delta t = 0.1$. The spatial resolution ensures that the mesh size close to the spheres is about $0.05 R$. 

\begin{figure}[t]
\begin{center}
\includegraphics[width=.3\textwidth,trim=0cm 0cm 3.3cm 0cm,clip]{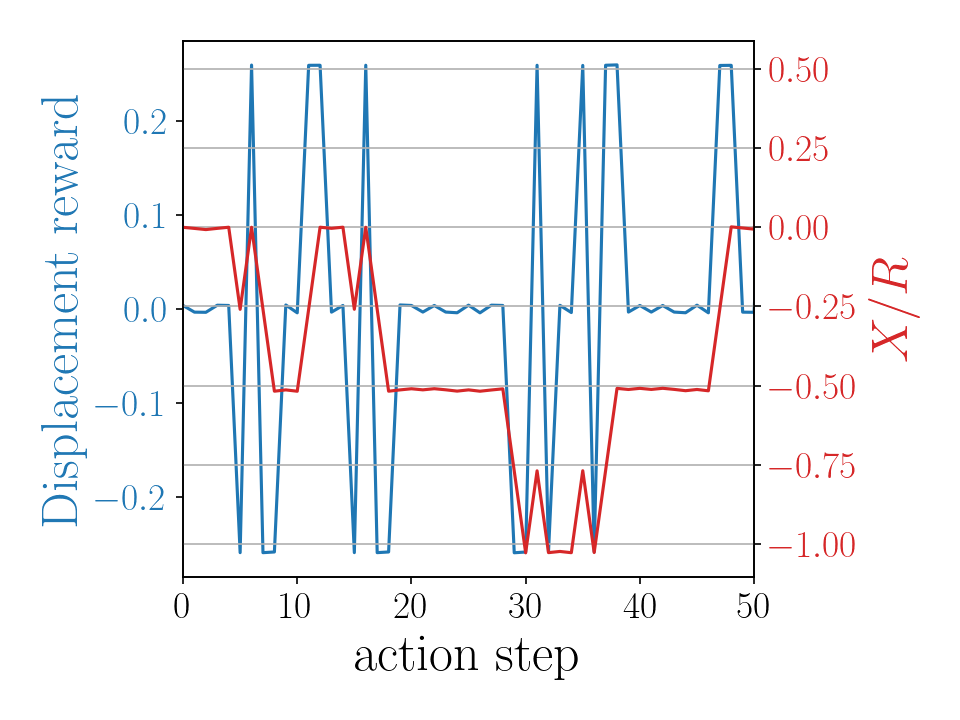}
\includegraphics[width=.3\textwidth,trim=0cm 0cm 3.3cm 0cm,clip]{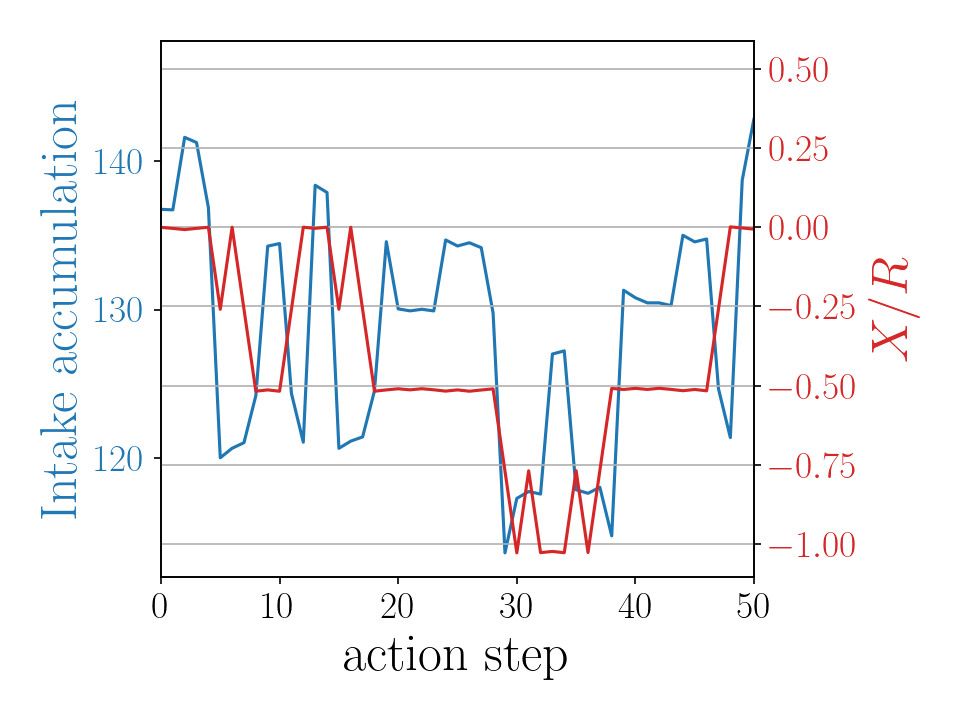}
\includegraphics[width=.38\textwidth]{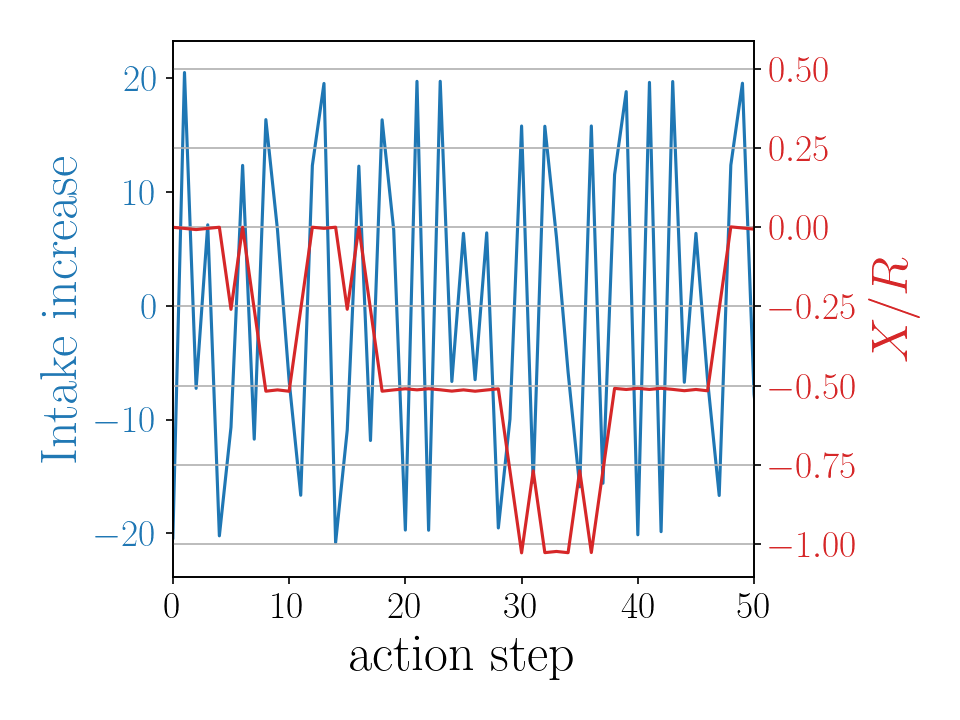}
\end{center}
\caption{Rewards (blue) and normalized center of mass position (red) for $50$ actions for a realization with Pe $ = 0.06$. The red vertical scale is the same for all cases.}\label{fig:rvvsxc}
\end{figure}

In this way, long sequences of actions were computed and stored for random motions at different P\'eclet numbers ranging from Pe $=0.06$, which is fully diffusion dominated, to Pe $=60$. The instantaneous rewards $r_t$ corresponding to the first $50$ actions of some of the recorded sequences are plotted in Fig.~\ref{fig:rvvsxc}, together with the position $X_t$. 

Let us first discuss the results obtained at Pe $=0.06$.
This is the case with lowest complexity, since the concentration field does not depend on the history and evolves quasi-statically. We divided the generated sequence of 6000 experiences into $12$ sets of $500$ consecutive experiences and ran the Q-learning process on each experience set for different values of $\alpha$ and $\gamma$. The resulting policy $\pi_{\text{\tiny QL}}$ was compared to $\pi^*$, classifying the learning process as successful if $\pi_{\text{\tiny QL}}=\pi^*$. 

\begin{figure}[t]
\begin{center}
\includegraphics[width=1\textwidth,trim=.3cm .4cm 0cm 0cm,clip]{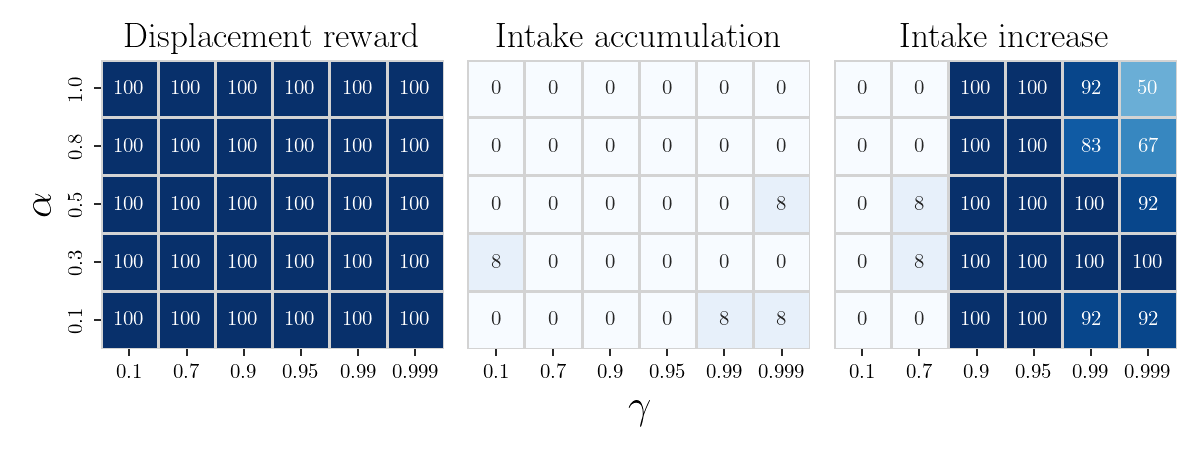}
\end{center}
\caption{Heat maps of percentage of learning success obtained from 12 sets of 500 consecutive experiences for the data set with Pe = 0.06.}\label{fig:bshm006long}
\end{figure}

\begin{figure}[th!]
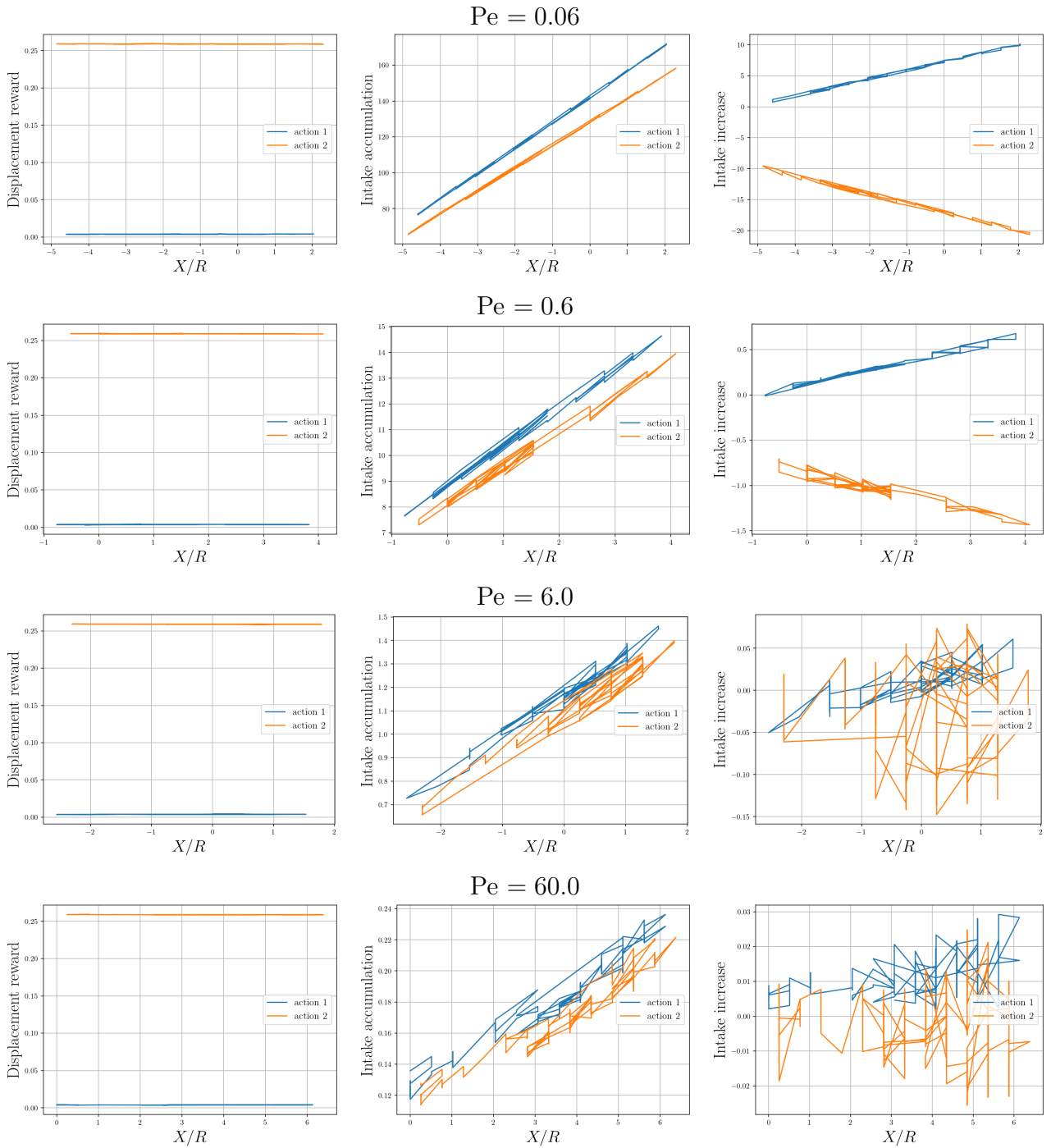

\begin{center}
\foreach \pp\ppt in {0.06/0.06,0.60/0.6,6.00/6.0,60.0/60.0}
{
Pe $ = \ppt$\par
\includegraphics[width=.325\textwidth]{rwvsxcPe\pp Tr1.0DispN.png}
\includegraphics[width=.325\textwidth]{rwvsxcPe\pp Tr1.0FAccN.png}
\includegraphics[width=.325\textwidth]{rwvsxcPe\pp Tr1.0FDifN.png}
}
\end{center}
\caption{Instantaneous rewards plotted as function of the center-of-mass position $X$ for different values of Pe. Only experiences that start at state $s_t=2$ are included, with different colors for the cases $a_t=1$ (blue) and $a_t=2$ (yellow).}\label{fig:rewvsxc}
\end{figure}

The heat maps in Fig.~\ref{fig:bshm006long} show the success rates obtained for each reward. For reward R1 (displacement), the success rate is $100 \%$ for all tested values of $\alpha$ and $\gamma$. For reward R2 (intake accumulation) the situation is quite the opposite, since the success rate is zero for almost every pair $(\gamma,\alpha)$. In the case of reward R3 (intake increase), values of $\gamma \geq 0.9$ show high success rates, but any $\gamma \leq 0.7$ fails consistently. This suggests that, even at low Pe, learning chemotaxis is more difficult than learning locomotion. This can be explained by looking at the top row of Figure \ref{fig:rewvsxc}, where the instantaneous rewards $r_{t+1}$ of actions $a=1$ and $a=2$ are plotted against $X_t$ for the state $s_t=2$. Reward R1 does not depend on $X$, it does not vary along the evolution. The other rewards are affine on $X_t$, which makes them evolve over time and increases the difficulty of learning.

The sharp constancy of $\Delta X$ for each action is an additional verification of the accuracy of the numerical model. Though the mesh is different each time the action is executed, $\Delta X$ does not change. 

Rewards R2 and R3 show some small noise on top of their affine dependence on $X_t$. This is an effect of the non-zero Péclet number. If we now look at the other rows of Figure~\ref{fig:rewvsxc}, we see that the variance of the reward for each action and each $X_t$ increases significantly with Pe, for both R2 and R3. Already at Pe $=6$ the rewards of the two actions overlap for the same $X_t$, making it even more difficult to predict the convenience of one action over the other. 

\begin{figure}[p]
\begin{center}
\includegraphics[width=1\textwidth]{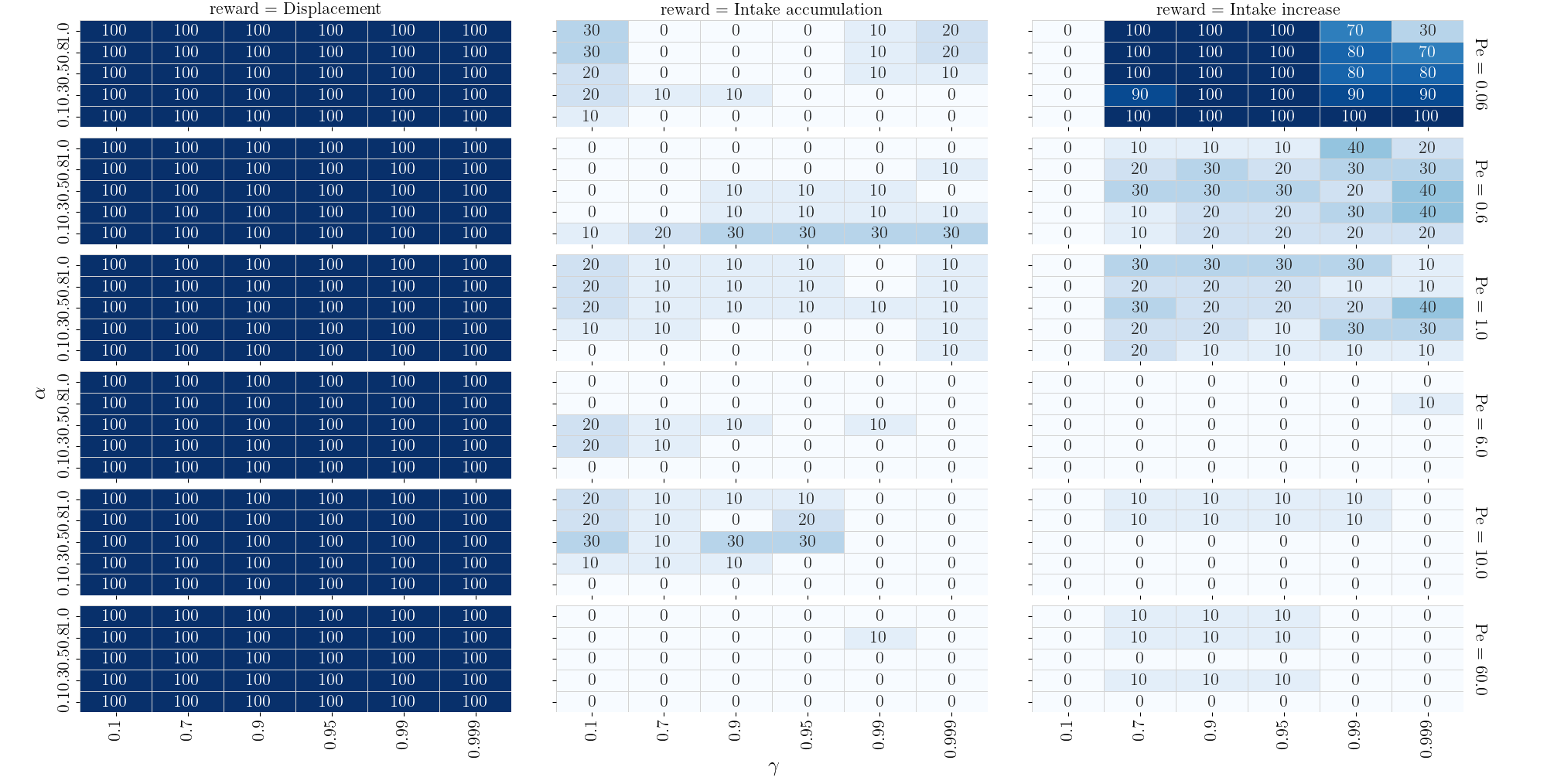}
\end{center}
\caption{Heat maps of percentage of learning success obtained from $10$ sets of $500$ consecutive experiences for the data sets corresponding, from top to bottom, to Pe = 0.06, 0.6, 1, 6, 10 and 60.}\label{fig:successratevspeclet500}
\end{figure}

The consequences on the success rate of Q-learning at chemotaxis are as would be expected. We extracted 10 batches of 500 consecutive experiences from the  the sequences of 1000 experiences collected at each Pe $\geq$  0.6. 
In Fig.~\ref{fig:successratevspeclet500} we show the success rates over the ten batches as heat maps over the parameters $\gamma$ and $\alpha$ as before.

\begin{figure}[p]
\begin{center}
\includegraphics[width=1\textwidth]{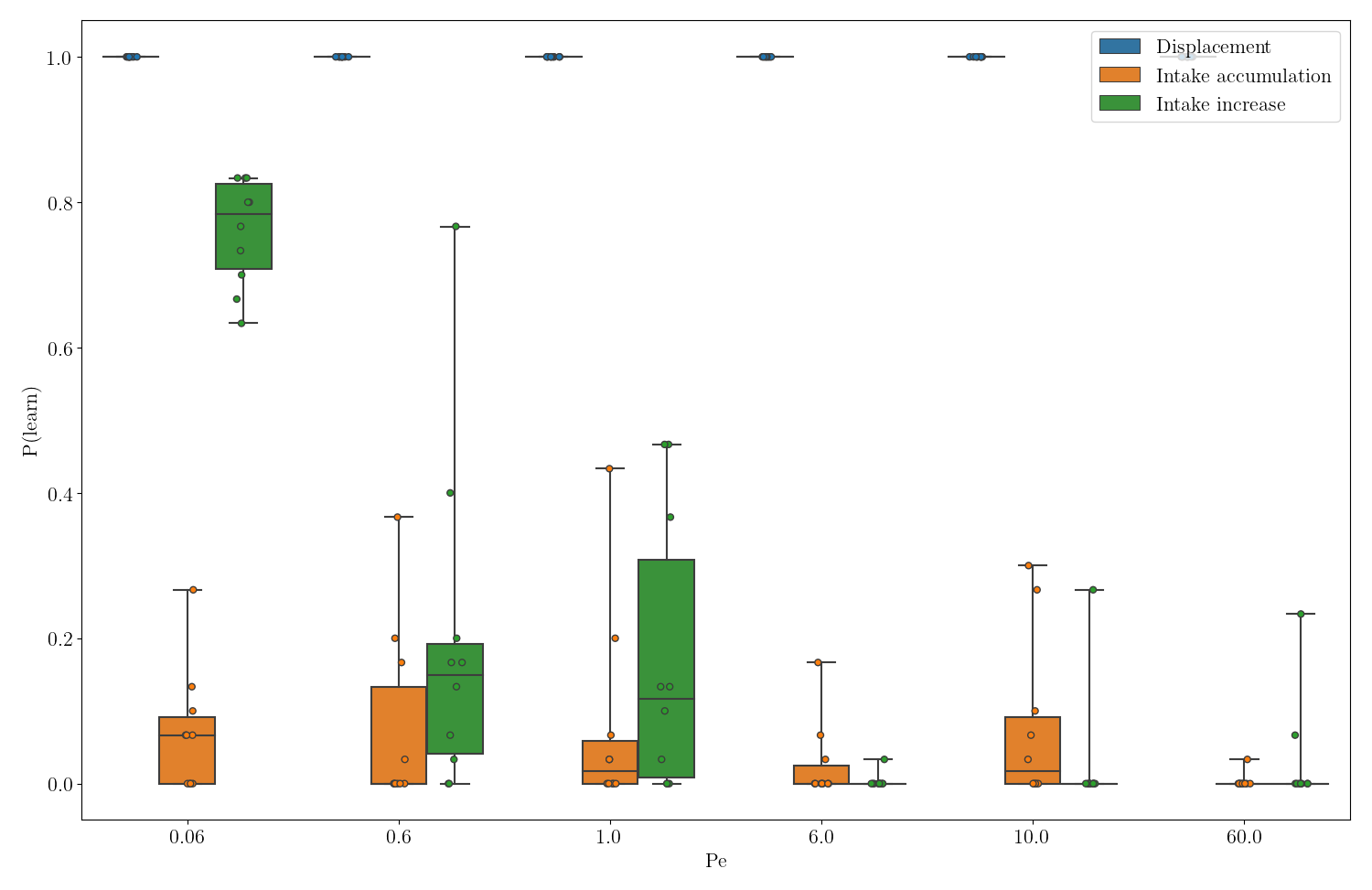}
\end{center}
\caption{Box plot of the total success rate (summed over $\gamma$ and $\alpha$) for each reward and different Péclet numbers. Each data point corresponds to one of ten batchs of 500 experiences each.}\label{fig:boxplot500}
\end{figure}

The displacement reward is unaffected by Pe and thus remains at 100 \% success rate throughout the maps. Reward R2, that performs very poorly at Pe = 0.06, improves somewhat for higher Pe, outperforming R3 at Pe  = 6 and 10. Reward R3 exhibits high success rates at low Pe but they decline systematically as Pe increases to 6. Some improvement is then observed at Pe = 60. These observations are summarized in Fig. \ref{fig:boxplot500}, which shows box plots of the total success rate (summed over $\gamma$ and $\alpha$) as function of Pe, aggregating the data corresponding to the ten learning batches of experiences. 

\section{Conclusions}

We have examined the coupled problem of hydrodynamics and solute transport for a three-sphere swimmer at P\'eclet numbers ranging from 0.06 to 60. The adopted method is the numerical simulation of the problem with a finite element code based upon the FEniCS library.

We first analyzed the swimming gait and produced previously unavailable data on the dependence of the Sherwood number with the Péclet number in homogeneous environment. They confirm that little gain in solute flux can be achieved by swimming (with respect to staying still in the fully extended position)  unless Pe is significantly larger than 10. 

We then turned to considering the swimmer as an agent moving inside a fluid that has a concentration gradient (chemotaxis). Inside a positive gradient, the swimmer/agent has potentially unbounded gain in solute flux if it swims toward the right. 

A Q-learning process was applied to sets of experiences collected along random decisions of the swimmer/agent to evaluate the difficulty of learning chemotaxis as compared to that of learning locomotion. The numerical evidence shows that the latter is significantly easier than the former. The chemotaxis problem even at low Pe has a varying environment that renders learning more difficult. Further, the average success rate falls sharply for Pe$\geq 1$ and the dispersion in the outcome increases.

The overall conclusion is that proprioception (information about $s_t$) and chemoreception (information about $J(t)$) alone pose a learning problem for chemotaxis that is quite hard. Several routes toward alleviating this difficulty exist. Further sensory information can be added, such as the fraction of the flux that is absorbed by each sphere. Information about the concentration field away from the swimmer could possibly be made available by antenna-like chemoreceptive sensors. Another route consists of considering that the swimmer/agent has an "innate" bias towards random motions different from the totally random motions used in this study. This would induce a preferential sampling of the space of sequences of actions that could improve learning. These routes, set for future work, will help elucidate the challenges that swimmers must overcome to learn chemotaxis.

\section{Acknowledgements}
The authors gratefully acknowledge the financial support received from the S\~ao Paulo Research Foundation FAPESP/\-CEPID/CeMEAI grant 2013/07375-0, and from Brazilian National Council for Scientific and Technological Development CNPq grants 305599/2017-8 and 310990/2019-0. This research was carried out using computational resources from the Cluster-Sala Jürgen Tischer, Department of Mathematics, Universidad del Valle, Cali, Colombia.
The authors are thankful to the developers of the free and open source software used for this publication.
\newpage
\bibliographystyle{unsrtnat}
\bibliography{references.bib}

\end{document}